\documentclass[10pt,twocolumn,letterpaper]{article}

\usepackage{ijcb}
\usepackage{times}
\usepackage{epsfig}
\usepackage{graphicx}
\usepackage{amsmath}
\usepackage{amssymb}
\usepackage{multirow}
\usepackage{balance}
\usepackage{float}

\ijcbfinalcopy 

\ifijcbfinal\fi
\begin{document}

\title{{D-NetPAD}: An Explainable and Interpretable Iris Presentation Attack Detector}

\author{Renu Sharma \hspace{2cm} Arun Ross\\
Department of Computer Science and Engineering\\
Michigan State University\\
{\tt\small \{sharma90, rossarun\}@cse.msu.edu}
}

\maketitle
\thispagestyle{empty}

\begin{abstract}
   An iris recognition system is vulnerable to presentation attacks, or PAs, where an adversary presents artifacts such as printed eyes, plastic eyes, or cosmetic contact lenses to circumvent the system. In this work, we propose an effective and robust iris PA detector called D-NetPAD based on the DenseNet convolutional neural network architecture. It demonstrates generalizability across PA artifacts, sensors and datasets. Experiments conducted on a proprietary dataset and a publicly available dataset (LivDet-2017) substantiate the effectiveness of the proposed method for iris PA detection. The proposed method results in a true detection rate of 98.58\% at a false detection rate of 0.2\% on the proprietary dataset and outperfoms state-of-the-art methods on the LivDet-2017 dataset. We visualize intermediate feature distributions and fixation heatmaps using t-SNE plots and Grad-CAM, respectively, in order to explain the performance of D-NetPAD. Further, we conduct a frequency analysis to explain the nature of features being extracted by the network. The source code and trained model are available at https://github.com/iPRoBe-lab/D-NetPAD.
\end{abstract}

\section{Introduction}
\begin{figure}[h!]
	\includegraphics[width=\linewidth]{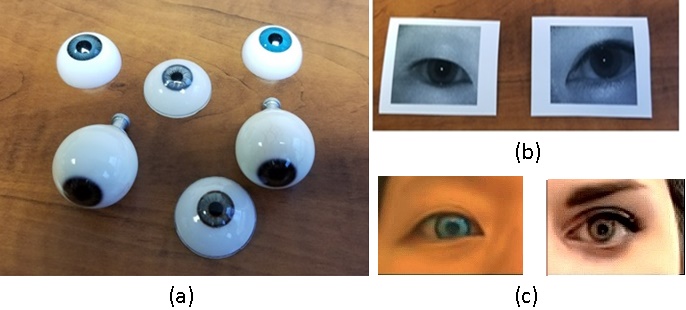}
	\caption{Example of presentation attacks (PAs) launch on the iris modality: (a) artificial eyes, (b) printed iris images, and (c) cosmetic contacts.}
	\label{fig:SamplePAs}
\end{figure}

An iris biometric system recognizes an individual based on the textural pattern of their iris \cite{Daugman1993}. The increasing popularity of iris systems and their unattended mode of operation make them susceptible to {\em presentation attacks}. According to ISO/IEC 30107-1:2016 \cite{ISO30107-1-2016}, a Presentation Attack (PA) is a ``presentation to the biometric data capture subsystem with the goal of interfering with the operation of
the biometric system". 
The biometric characteristics or materials used to launch a presentation attack are termed as Presentation Attack Instruments (PAIs). Examples of PAIs in the case of the iris modality include printed iris images \cite{Daugman1999, Czajka2013, Raghavendra2015, Hoffman2018}, artificial eye (plastic, glass, or doll eyes) \cite{Hoffman2018, Lee2006}, cosmetic contacts \cite{Raghavendra2017, Yadav2014, Hughes2013}, video display of an eye image \cite{Raja2015, Czajka2018}, cadaver eyes \cite{Marcel2019, Czajka2018}, robotic eye models \cite{Komogortsev2015} and holographic eye images \cite{Pacut2006}. A few samples of iris PAIs are shown in Fig. \ref{fig:SamplePAs}. There is a need to detect these known iris PAs as well as other ``unknown" and ``unseen" PAs that may compromise the security of iris-based biometric systems.\footnote{An ``unknown" or ``unseen" attack involves using PAIs that were {\em not} observed in the training data.} \textit{In this work, our objective is to develop an effective and explainable iris PA detector.}

\begin{figure*}[h!]
	\includegraphics[width=\linewidth]{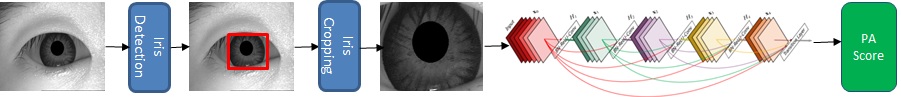}
	\caption{Flowchart of the D-NetPAD algorithm. Iris region (red box) is detected and cropped from the ocular image and input to the D-NetPAD architecture. The base architecture used in D-NetPAD is Dense121 \cite{Huang2017}. It produces a single PA score, which determines whether an input image is a bonafide or a PA.}
	\label{fig:D-NetPAD}
\end{figure*}

Existing techniques in the literature used to counter iris PAs can be categorized as being either hardware-based or software-based. Hardware-based techniques require physical devices in addition to the conventional iris sensor to aid in PA detection. Examples include the use of IrisCUBE camera to capture pupil dynamics \cite{Czajka2015}, 3D structural modeling of an eye using stereo imaging \cite{Hughes2013}, use of CCD camera with two white LEDs to initiate and record pupillary reflex \cite{Kanematsu2007} and EyeLink II eye tracker to capture Oculomotor Plant Characteristics \cite{Komogortsev2013}. These techniques incur an additional cost due to the hardware involved. Moreover, image acquisition using these methods is typically time-consuming and requires explicit user cooperation.  

On the other hand, software-based techniques extract salient features from the {\em digital} iris image in order to classify it as a bonafide or a PA.\footnote{A ``bonafide" image is sometimes referred to as a ``live" image in the literature.} These features can be either hand-crafted or can be learned using deep learning schemes. Examples of hand-crafted features used to detect iris PAs include SIFT \cite{Zhang2015}, LBP \cite{Hu2016}, BSIF \cite{Ramachandra2015}, and SID \cite{Gragnaniello2015}. However, more recently, a number of deep-learning based methods have been proposed \cite{Menotti2015,Pala2017,Chen2018a,Hoffman2019,DYadav2019,Yadav2020}. Menotti \textit{et al.} \cite{Menotti2015} propose a deep architecture for PA detection called SpoofNet. Pala and Bhanu \cite{Pala2017} develop a deep framework built upon triplet convolutional networks. Hoffman \textit{et al.} \cite{Hoffman2018} focus on detecting iris PAs utilizing a patch-batch convolutional neural network (CNN) that is observed to perform well in the cross-sensor and cross-dataset scenarios. They extend their work \cite{Hoffman2019} by analyzing the importance of utilizing the periocular region in detecting iris PAs. Chen and Ross \cite{Chen2018a} propose a multi-task CNN for first detecting the iris region and then classifying it. Yadav \textit{et al.} \cite{Yadav2020} utilize a Relativistic Average Standard Generative Adversarial Network (RaSGAN) as a one-class classifier to detect unseen or unknown iris PAs. The Liveness Detection-Iris Competition (LivDet-Iris) held in 2013 \cite{LivDet2013}, 2015 \cite{LivDet2015} and 2017 \cite{LivDet2017} provides a comprehensive comparative report of different iris PA detection techniques. Czajka and Bowyer \cite{Czajka2018} also present a detailed assessment of various state-of-the-art iris PA detection (PAD) algorithms. While most of these methods resulted in very high PA detection rates, generalizability across PAs, sensors, and datasets is still a challenging problem \cite{Yadav2014, Doyle2015, Raghavendra2017}. 

In this paper, we propose a CNN-based iris PAD method that utilizes the DenseNet \cite{Huang2017} architecture. Yadav \textit{et al.} \cite{DYadav2019} also utilize the DenseNet architecture to detect cosmetic contact PA images captured by various mobile iris sensors. On the other hand, our work considers a much large range of iris PAs captured by various desktop and mobile iris sensors. The DenseNet architecture has a unique property that each layer is connected to every other layer in a feed-forward fashion. The features across different layers correspond to different resolutions. The aggregated effect of multi-resolution features efficiently characterize the iris pattern as the iris pattern is highly stochastic in nature and the intricate features of the iris stroma are manifested in multiple resolution \cite{Daugman2001}. 
The main contributions of the work are as follows:
\begin{enumerate}
	\item We propose an effective and robust iris PA detector named as D-NetPAD that is based on the DenseNet architecture. We also demonstrate that the proposed detector exhibits generalizability across different PAs, sensors, datasets.
	\item We evaluate the performance of D-NetPAD on a proprietary dataset (Combined) as well as a publicly available dataset (LivDet-2017). 
	\item We perform visualizations using t-SNE plots \cite{Maaten2008} and Grad-CAM \cite{Selvaraju2017} to explain the performance of the proposed method. The t-SNE plots provide visualization of features obtained from the intermediate layers of the model. The Grad-CAM produces heatmaps emphasizing the salient regions in an iris image that are used by the network to detect iris PAs. 
	\item We also conduct a frequency analysis to understand the frequencies learned by the model and, based on that, interpret its performance.  
\end{enumerate}

Section 2 discusses the architecture of the proposed method. Section 3 describes the experimental setup and results on both the datasets. Section 4 provides a detailed analysis of the results obtained from the D-NetPAD. Finally, section 5 concludes the paper.

\section{D-NetPAD Description}

Dense Network Presentation Attack Detection (D-NetPAD) is based on the Densely Connected Convolutional Network 121 (DenseNet121) \cite{Huang2017} architecture. The architecture consists of 121 convolutional layers of kernel size 7 $\times$ 7, followed by a max-pooling layer and a series of Dense blocks and Transition layers. There are four Dense blocks, and three Transition layers lie between successive Dense blocks. Each Dense block consists of two convolutional layers of kernel size 1 $\times$ 1 and 3 $\times$ 3. Both convolutional layers are followed by a non-linear ReLU activation layer. The Transition layer consists of one convolutional layer of kernel size 1 $\times$ 1 and an average pooling layer. It reduces the size of feature-maps, which is kept constant within a Dense block. The last layer is a fully connected layer. The work in \cite{DYadav2019} exploits the DenseNet architecture of depth 22 with three densely connected blocks. 

The most notable characteristic of DenseNet is that each layer connects to every other layer in a feed-forward fashion. In other words, each layer obtains feature-maps from preceding layers and passes its feature-maps to subsequent layers. The features from preceding layers are combined by concatenation as opposed to the summation performed in the ResNet \cite{He2016} architecture. The concatenation removes the constraint of having the same dimension on the feature-maps. In this way, the architecture ensures the maximum flow of information in the forward direction and also resolves the most prevalent challenge of vanishing gradient in the backward direction. Another major advantage of DenseNet121 is that it supports such densely and deeply connected network with fewer trainable parameters (7,978,856) as compared to its counterpart ResNet50 (35,610,216) \cite{He2016} or VGG19 (143,667,240) \cite{Simonyan2015}. This is because DenseNet uses a small set of filters in each layer (e.g., 12 filters/layer) compared to the traditional convolutional networks ($\sim$128 or 256 filters/layer). DenseNet preserves the feature-maps and reuses it in the subsequent layers instead of relearning feature-maps every time. The reusability of feature-maps helps in alleviating the over-fitting problem, especially in the case of limited training data. These architectural tweaks help in generating an efficient feature representation for the highly textured iris pattern. Feature-maps at each layer capture specific spatial and frequency information and consolidation of these feature-maps result in the extraction of multi-resolution features. These features are efficient in characterizing the stochastic nature of the iris pattern. The intricacy of a bonafide iris pattern is not present in the spoofed iris (print eye, artificial eye, or cosmetic contact), and this difference is efficiently captured by the features generated from DenseNet. The consolidation of feature-maps at the last layer also smoothens the decision boundaries, resulting in better generalization across PA artifacts, sensors, and datasets.

Figure \ref{fig:D-NetPAD} shows the flowchart of the proposed architecture. The iris sensor acquires an ocular image which input to the iris detection module. In our implementation, we use the VeriEye iris detector, which outputs the centers of the iris and pupil along with their radii. The iris region is cropped from the ocular image using the center and radius of the iris. The cropped iris region is then resized to 224 $\times$ 224 and input to the pre-trained Dense121 network. The ImageNet dataset \cite{Deng2009} is used to pre-train the network. It produces a single presentation attack (PA) score, which lies between 0 and 1. A score approaching `1' indicates that the input sample is a PA, whereas a score approaching `0' indicates that the input sample is a bonafide. We determine the threshold by fixing the False Detection Rate to 0.2\% in order to get the final classification. If the PA score is less than the specified threshold, the input sample is labeled as a bonafide; otherwise, it is a PA. During training, the learning rate used is 0.005, the batch size is 20, the optimization algorithm used is the stochastic gradient descent with a momentum of 0.9, the number of epochs is 50, and the loss function is cross-entropy.

\begin{figure*}[h!]
	\includegraphics[width=\linewidth]{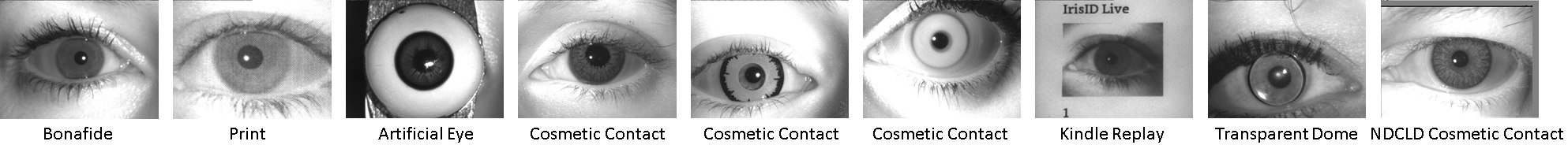}
	\caption{Sample images of bonafide, different types of PAs (print, artificial eye, cosmetic contact, kindle replay, and transparent dome on print) taken from the Combined dataset. The last cosmetic contact image is taken from the NDCLD-2015 dataset.}
	\label{fig:JHU-Images}
\end{figure*}

\section{Evaluation and Results }
We performed experiments on a proprietary dataset and a publicly available benchmark dataset (LivDet-2017) to evaluate the performance of D-NetPAD. The proprietary dataset has several subsets and is, therefore, referred to as the ``Combined Dataset" in the rest of the document. The Combined dataset corresponds to the cross-PA scenario, whereas the LivDet-2017 dataset creates a test-bed for cross-PA, cross-sensor, and cross-dataset testing scenarios. In the cross-PA scenario, we use PA instruments (PAIs) that were not used during the training. In the cross-sensor scenario, we evaluate images from different sensors than those used during the training. The cross-dataset scenario incorporates testing under different PAIs, sensors, data acquisition environments (indoor/outdoor, varying illumination conditions), subject populations, and platforms (desktop or mobile). The cross-dataset scenario accounts for large variations, making it the most challenging test scenario.

\subsection{Combined Dataset: Description and Results}
The Combined Dataset was collected under the IARPA Odin program (Presentation Attack Detection) \cite{Odin}. The IrisAccess iCAM7000 sensor was used to collect the data. The dataset is a combination of various component datasets collected at different locations and times using different units of the same sensor. Table \ref{table:JHU-Dataset} provides the description of the component datasets. There are a total of 13,851 iris images out of which 9,660 are bonafide and 4,291 are PAs. The PA samples in the dataset correspond to the following attack instruments: print, artificial eye, cosmetic contacts, kindle replay, and transparent dome on print. Figure \ref{fig:JHU-Images} shows sample images from the dataset. The test set JHU-APL03 (Table \ref{table:JHU-Dataset}) comprises two types of artificial eyes and 10 different types of cosmetic contacts. It corresponds to the cross-PA scenario as it contains six additional cosmetic contacts that are not used during training. As the process of collecting cosmetic contact images is a tedious and time-consuming process, its quantity is limited in the training set. Therefore, we utilize cosmetic contact images from NDCLD-2015 \cite{NDCLD2015} to overcome the shortcoming. The bonafide images in the NDCLD-2015 dataset are not used as the Combined dataset has a large number of bonafide images. The NDCLD-2015 dataset was collected using the IrisGuard AD100 and IrisAccess LG4000 sensors.

\begin{table*}[h!]
	\caption{Description of different components of the Combined Dataset. Details of the train and test set of the Combined and NDCLD 2015 datasets are also provided in terms of the number of bonafide and PA images. AA and BB are used for anonymization. JHU-APL stands for Johns Hopkins University-Applied Physics Laboratory.}
	\label{table:JHU-Dataset}
	\centering
	\resizebox{\linewidth}{!}{%
	\begin{tabular}{|l|c|c|c|c|c|c|c|c|c|c|}
		\hline
		\multirow{2}{*}{\textbf{Dataset}} & \multicolumn{9}{c|}{\textbf{Train}} & \textbf{Test} \\ \cline{2-11} 
		& AA IrisPA01 & BB IrisPA01 & BB IrisPA02 & SelfTest01 & SelfTest02 & SelfTest03 & JHU-APL01 & JHU-APL02 & NDCLD 2015 & JHU-APL03 \\ \hline
		Bonafide & 381 & 962 & 1,107 & 446 & 518 & 518 & 1,394 & 1,371 & - & 2,963 \\ \hline
		Print & 991 & 660 & 415 & 14 & - & - & - & - & - & - \\ \hline
		Artificial Eye & 318 & 34 & - & 21 & 9 & 12 & 49 & 111 & - & 175 \\ \hline
		Cosmetic Contacts & - & - & 208 & - & 21 & 94 & 78 & 120 & 2,236 & 177 \\ \hline
		Kindle Replay & 51 & 79 & - & - & - & - & - & - & - & - \\ \hline
		Transparent Dome & - & - & 503 & 9 & - & - & 42 & - & - & - \\ \hline
		Acquisition Time Period & \multicolumn{1}{l|}{Nov 2017} & \multicolumn{1}{l|}{Nov 2017} & \multicolumn{1}{l|}{Dec 2018} & \multicolumn{1}{l|}{April 2018} & \multicolumn{1}{l|}{Feb 2019} & \multicolumn{1}{l|}{Sept 2019} & \multicolumn{1}{l|}{May 2018} & \multicolumn{1}{l|}{May 2019} & 2015 & \multicolumn{1}{l|}{Nov 2019} \\ \hline
	\end{tabular}
}
\end{table*}

\begin{figure}[h!]
	\centering
	\includegraphics[width=0.8\linewidth]{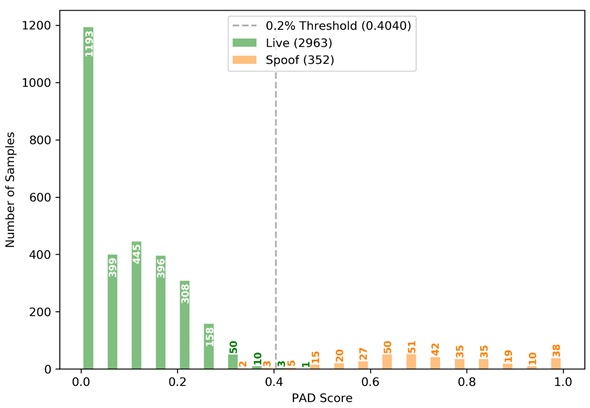}
	\caption{The histogram of PA scores corresponds to bonafide (green) and PAs (orange) generated by the D-NetPAD on the JHU-APL03 test set. The dotted vertical line represents the threshold (0.40) selected by setting the FDR at 0.2\%. At the specified threshold, 4 bonafide and 5 PA images are misclassified.}
	\label{fig:Histogram-D-NetPAD}
\end{figure}

\begin{figure}[h!]
	\includegraphics[width=\linewidth]{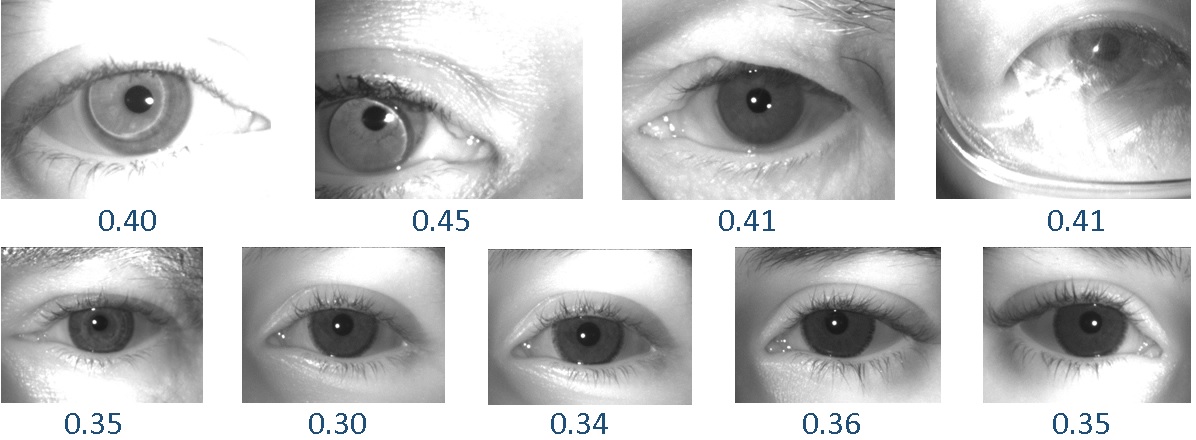}
	\caption{Misclassified iris images by the D-NetPAD algorithm on the JHU-APL03 test set. The first row shows bonafide images, which are misclassified as PA. The second row shows PA images, which are misclassified as bonafide. The PA score of the images is displayed at the bottom of each image. The threshold for classification is 0.40, where a PA score below the threshold is considered to be a bonafide.} 
	\label{fig:JHU-Missclassified}
\end{figure}

We evaluate the performance of the D-NetPAD in terms of True Detection Rate (TDR) at a False Detection Rate (FDR) of 0.2\%. TDR is the percentage of PA samples correctly detected, whereas FDR is a percentage of bonafide samples incorrectly classified as PA.\footnote{Other commonly used evaluation measures for presentation attack detection are Attack Presentation Classification Error Rate (APCER) and Bonafide Presentation Classification Error Rate (BPCER). TDR is 1$-$APCER, and FDR is the same as BPCER.} The D-NetPAD is compared against two deep learning-based methods (\cite{Chen2018a} and \cite{Hoffman2019}) as these are state-of-the-art (SoTA) methods. It is also compared with VGG19 \cite{Simonyan2015} and ResNet101 \cite{He2016} deep architectures. Table \ref{table:JHU-Results} presents the results of all five algorithms. The D-NetPAD outperforms the SoTA methods \cite{Hoffman2019}, \cite{Chen2018a}, VGG19 and ResNet101 by 25.27\%, 6.44\%, 2.41\% and 1.75\%, respectively. It resulted in 98.58\% TDR at 0.2\% FDR. The performance of D-NetPAD is further analyzed using the histogram of PA scores shown in Figure \ref{fig:Histogram-D-NetPAD}. The vertical line in the figure represents the threshold corresponding to 0.2\% FDR. At the selected threshold, four bonafide samples are misclassified as PAs and five PAs are misclassified as bonafides. Figure \ref{fig:JHU-Missclassified} shows the misclassified images. In the case of misclassified bonafide images, subjects in the first two images were wearing hard transparent contact lenses that closely resemble cosmetic contact lenses. The subject in the third image wore a soft transparent lens, which may also have been confused with a cosmetic contact lens. The last image contains glare of the light reflected from the glasses, resulting in a misclassification. In the case of misclassified PA images, the D-NetPAD fails for a particular type of cosmetic contact lens (Halloween-style Extreme contact lens), where the pattern appears only at the periphery of the cosmetic contact. 
Segmentation ignores the outer region of the iris containing the artifacts of these cosmetic contacts. This resulted in a smaller region of the artifact being fed into the DenseNet for PA detection, leading to a misclassification.

\begin{table}[h!]
	\caption{The results of D-NetPAD in term of TDR (\%) at 0.2\% FDR on the Combined dataset. The method is compared with two other algorithms that have participated in the Odin program of IARPA.}
	\label{table:JHU-Results}
	\centering
	\resizebox{\columnwidth}{!}{%
		\begin{tabular}{|l|l|l|l|l|l|}
			\hline
			Algorithms & \cite{Hoffman2019} & \cite{Chen2018a} & VGG19 & ResNet101 & D-NetPAD \\ \hline
			TDR (\%) @ 0.2\% FDR & 78.69 & 92.61 & 96.26 & 96.88 & \textbf{98.58} \\ \hline
		\end{tabular}
	}
\end{table}

\subsection{LivDet-2017 Dataset: Description and Results}
Another dataset used for evaluation was the LivDet-2017 \cite{LivDet2017} dataset. The LivDet-2017 dataset is a combination of four datasets: Clarkson, Warsaw, Notre Dame, and IIITD-WVU datasets. Table \ref{table:LivDet-Dataset} describes the types of PAs present in the datasets, and the number of images in the train and test sets of all four datasets. The Clarkson dataset represents the cross-PA testing scenario. The test set consists of 5 additional cosmetic contacts and prints of visible spectrum iris images captured using an iPhone 5. The Warsaw dataset helps in evaluating the cross-sensor testing scenario. It consists of two test sets: a ``known" sensor and an ``unknown" sensor. The IrisGuard AD100 sensor is used to capture the images of the training set and the known ``known" component of the test set. Images of the ``unknown" component of the test set are captured by a setup composed of Aritech ARX-3M3C camera, SONY EX-View CCD sensor, Fujinon DV10X7.5A-SA2 lens, and B+W 092 NIR filter. The Notre Dame dataset corresponds to the cross-PA scenario. It also contains two test sets (``known" and ``unknown"). The ``unkown" test set includes cosmetic contacts not used in the training set. The IIITD-WVU dataset consists of data collected by IIITD and WVU. The IIITD data is used for training, whereas the WVU data is used for testing. The dataset corresponds to the cross-dataset scenario, where the test set incorporates variations in the sensors, data acquisition environment, subject population, and PA generation procedures. The training set is captured in a controlled environment using two iris sensors: Cogent dual iris sensor (CIS 202) and VistaFA2E single iris sensor. The test set is captured using the IriShield MK2120U mobile iris sensor at two different locations: indoors (controlled illumination) and outdoors (varying environmental conditions). The cross-dataset testing scenario represents the most difficult case.

\begin{table*}[h!]
	\caption{Description of the train and test sets of all four subsets of the LivDet-2017 dataset along with the number of bonafide and PA images present in the datasets. The information about the sensors is also provided. Each subset represents different testing scenarios. The Clarkson and Notre Dame test sets correspond to the cross-PA scenario, whereas the Warsaw data corresponds to the cross-sensor scenario. The IIITD-WVU represents a cross-dataset scenario. Here, ``K. Test" means a known test set of the dataset, and ``U. Test" means an unknown test set.}
	\label{table:LivDet-Dataset}
	\resizebox{\textwidth}{!}{%
		\begin{tabular}{|l|l|l|l|l|l|l|l|l|l|l|}
			\hline
			\multirow{2}{*}{\textbf{Dataset}} & \multicolumn{2}{c|}{\textbf{\begin{tabular}[c]{@{}c@{}}Clarkson\\ (Cross-PA)\end{tabular}}} & \multicolumn{3}{c|}{\textbf{\begin{tabular}[c]{@{}c@{}}Warsaw\\ (Cross-Sensor)\end{tabular}}} & \multicolumn{3}{c|}{\textbf{\begin{tabular}[c]{@{}c@{}}Notre Dame\\ (Cross-PA)\end{tabular}}} & \multicolumn{2}{c|}{\textbf{\begin{tabular}[c]{@{}c@{}}IIITD-WVU\\ (Cross-Dataset)\end{tabular}}} \\ \cline{2-11} 
			& Train & Test & Train & K. Test & U. Test & Train & K. Test & U. Test & Train & Test \\ \hline
			Bonafide & 2,469 & 1,485 & 1,844 & 974 & 2,350 & 600 & 900 & 900 & 2,250 & 702 \\ \hline
			Print & 1,346 & 908 & 2,669 & 2,016 & 2,160 & \multicolumn{1}{c|}{-} & \multicolumn{1}{c|}{-} & \multicolumn{1}{c|}{-} & 3,000 & 2,806 \\ \hline
			Cosmetic Contacts & 1,122 & 765 & \multicolumn{1}{c|}{-} & \multicolumn{1}{c|}{-} & \multicolumn{1}{c|}{-} & 600 & 900 & 900 & 1,000 & 701 \\ \hline
			Sensor & \multicolumn{2}{l|}{\begin{tabular}[c]{@{}l@{}}IrisAccess\\ EOU2200\end{tabular}} & \multicolumn{2}{l|}{\begin{tabular}[c]{@{}l@{}}IrisGuard\\ AD100\end{tabular}} & \begin{tabular}[c]{@{}l@{}}Aritech ARX-3M3C,\\ Fujinon DV10X7.5A,\\ DV10X7.5A-SA2 lens\\ B+W 092 NIR filter\end{tabular} & \multicolumn{3}{l|}{\begin{tabular}[c]{@{}l@{}}IrisGuard AD100,\\ IrisAccess LG4000\end{tabular}} & \begin{tabular}[c]{@{}l@{}}Cogent \\ CIS 202,\\ VistaFA2E\end{tabular} & \begin{tabular}[c]{@{}l@{}}IriShield \\ MK2120U\end{tabular} \\ \hline
	\end{tabular}}
\end{table*}


For the detailed evaluation of the D-NetPAD, we created three models of the D-NetPAD network, which differ in their training process:(i) \textbf{Pre-trained D-NetPAD}: The model trained on the Combined dataset is directly used; (ii) \textbf{Scratch D-NetPAD}: The model is trained from scratch on the LivDet-2017 train sets; and (iii) \textbf{Fine-tuned D-NetPAD}: The model that is pre-trained on the Combined dataset is fine-tuned on the LivDet-2017 train sets. The performance measure used is the same as used in \cite{LivDet2017}: Attack Presentation Classification Error Rate (APCER) and Bonafide Presentation Classification Error Rate (BPCER). The APCER is the proportion of PA samples misclassified as bonafide, whereas the BPCER is a proportion of bonafide samples misclassified as PAs. The D-NetPAD is compared against the top three winners of the LivDet-2017 competition. Table \ref{table:LivDet-Results} summarizes the results of all algorithms. While the pre-trained D-NetPAD model and the model trained from scratch perform at par with the state-of-the-art methods, the fine-tuned D-NetPAD model outperforms the other methods. 

We also measured the performance of D-NetPAD in terms of its TDR at 0.2\% FDR on the LivDet-2017 dataset. Table \ref{table:LivDet-Results-TDR} compiles the results of D-NetPAD on all four datasets of the LivDet-2017 [1] dataset. A summary of the results is provided below:

\textbf{Clarkson Test Dataset}: The pre-trained D-NetPAD fails on the test set of Clarkson. The Clarkson dataset corresponds to the cross-sensor and cross-PA scenarios. The images captured from IrisAccess EOU2200 is visually quite different from the images captured by the iCAM 7000 iris sensor, which results in the poor performance (28.63\%). But, the result improves (92.05\% and 93.51\%) when the training set (scratch or fine-tuned) includes the Clarkson train set (sensor information).

\textbf{Warsaw Test Dataset}: The pre-trained D-NetPAD achieves competent performance on the Warsaw dataset. The sensors and types of PA used in the Warsaw dataset are different from the one used in the training, but the images captured by the test sensors are visually similar, which results in comparable TDR. Fine-tuning the pre-trained D-NetPAD using the train set of Warsaw dataset results in 100\% TDR.

\textbf{Notre Dame Test Dataset}: The dataset represents the cross-PA scenario, where the test set uses additional cosmetic contacts. The pre-trained D-NetPAD model trained on diverse cosmetic contacts generalizes well across previously unseen cosmetic contacts (93.55\% and 91\%). Its performance drops on the unknown test set (66.55\%) when the model is trained from scratch as the diversity of cosmetic contacts is limited in the Notre Dame train set. Fine-tuning the model with the Notre Dame train set achieves 100\% TDR.

\begin{figure}[h!]
	\includegraphics[width=\linewidth]{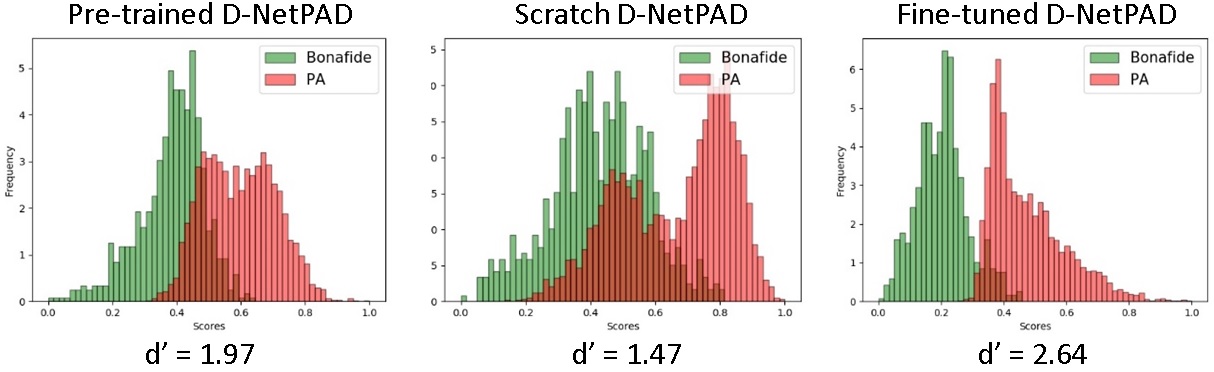}
	\caption{Histograms of the three trained models of D-NetPAD on the IIITD-WVU test set. For accurate classification, there should be minimal overlap between the two (red and green) distributions. This plot indicates the efficacy of the fine-tuned D-NetPAD.}
	\label{fig:Histogram-IIITD-WVU}
\end{figure}

\textbf{IIIT-WVU Test Dataset}: The dataset is the most challenging dataset where the test set images are captured using the IriShield MK2120U mobile iris sensor and under different capturing environment (indoor and outdoor). The dataset also included unseen PAs, resulting in very low TDRs for all three models (42.91\%, 29.30\%, and 48.85\%). We further analyze the results of IIIT-WVU by plotting the PA score distributions of the bonafide and PAs, and estimating the d-prime distance between them (Figure \ref{fig:Histogram-IIITD-WVU}). Though the TDR is quite low in the case of fine-tuned D-NetPAD, its histogram shows a better separation (d$^\prime=2.64$) between the score distributions of bonafide and PAs. 

The D-NetPAD algorithm demonstrates robustness across PAs and sensors testing scenarios after the fine-tuning but fails in the case of cross-dataset which is a combination of cross-PA, cross-sensor, cross-environment, and cross-platform scenarios. Here, cross-platform implies training on images of iris sensor meant for desktop (e.g., IrisAccess iCAM7000) and testing on images of iris sensor meant for mobile devices (e.g., IriShield MK2120U).

\begin{table*}[h!]
	\caption{D-NetPAD performance reported in terms of APCER and BPCER on all subsets of the LivDet-2017 dataset. The method is compared with three state-of-the-art algorithms in \cite{LivDet2017}, which are the winners of the LivDet 2017 competition.}
	\label{table:LivDet-Results}
	\resizebox{\textwidth}{!}{%
		\begin{tabular}{|l|l|l|l|l|l|l|l|l|l|l|}
			\hline
			\multirow{2}{*}{Algorithm} & \multicolumn{2}{l|}{Clarkson} & \multicolumn{2}{l|}{Warsaw} & \multicolumn{2}{l|}{IIITD-WVU} & \multicolumn{2}{l|}{Notre-Dame} & \multicolumn{2}{l|}{Averaged} \\ \cline{2-11} 
			& APCER & BPCER & APCER & BPCER & APCER & BPCER & APCER & BPCER & APCER & BPCER \\ \hline
			CASIA & 9.61 & 5.65 & 3.4 & 8.6 & 23.16 & 16.1 & 11.33 & 7.56 & 11.88 & 9.48 \\ \hline
			Anon1 & 15.54 & 3.64 & 6.11 & 5.51 & 29.4 & \textbf{3.99} & 7.78 & 0.28 & 14.71 & 3.36 \\ \hline
			UNINA & 13.39 & \textbf{0.81} & 0.05 & 14.77 & 23.18 & 35.75 & 25.44 & 0.33 & 15.52 & 12.92 \\ \hline
			Pre-Trained D-NetPAD & 16.73 & 19.46 & 1.66 & 0.83 & 16.05 & 15.24 & 1.00 & 2.22 & 8.86 & 9.43 \\ \hline
			Scratch D-NetPAD & 5.78 & 0.94 & \textbf{0} & \textbf{0.04} & 36.41 & 10.12 & 10.38 & 3.23 & 13.14 & 3.58 \\ \hline
			Fine-tuned D-NetPAD & \textbf{2.99} & 2.97 & \textbf{0} & 0.54 & \textbf{1.88} & 8.84 & \textbf{0.33} & \textbf{0.27} & \textbf{1.3} & \textbf{3.15} \\ \hline
	\end{tabular}}
\end{table*}

\begin{table}[h!]
	\caption{D-NetPAD performance reported in terms of the TDR (\%) @ 0.2\% FDR on different subsets of the LivDet-2017 dataset. Three models of D-NetPAD are generated by varying their training data.}
	\label{table:LivDet-Results-TDR}
	\resizebox{\columnwidth}{!}{
		\begin{tabular}{|l|c|l|c|c|c|c|c|}
			\hline
			\multirow{2}{*}{Algorithm} & \multicolumn{2}{c|}{Clarkson} & \multicolumn{2}{c|}{Warsaw} & \multicolumn{2}{c|}{Notre-Dame} & IIITD-WVU \\ \cline{2-8} 
			& \multicolumn{2}{c|}{Test} & K. Test & U. Test & K. Test & U. Test & Test \\ \hline
			Pre-Trained D-NetPAD & \multicolumn{2}{c|}{28.63} & 92.95 & 98.56 & 93.55 & 91.00 & 42.91 \\ \hline
			Scratch D-NetPAD & \multicolumn{2}{c|}{92.05} & 100 & 100 & 100 & 66.55 & 29.30 \\ \hline
			Fine-tuned D-NetPAD & \multicolumn{2}{c|}{93.51} & 100 & 100 & 100 & 99.77 & 48.85 \\ \hline
		\end{tabular}
	}
\end{table}

\section{Explainability analysis}

\subsection{Visualization Analysis}

\begin{figure*}[h!]
	\centering
	\includegraphics[width=0.85\linewidth]{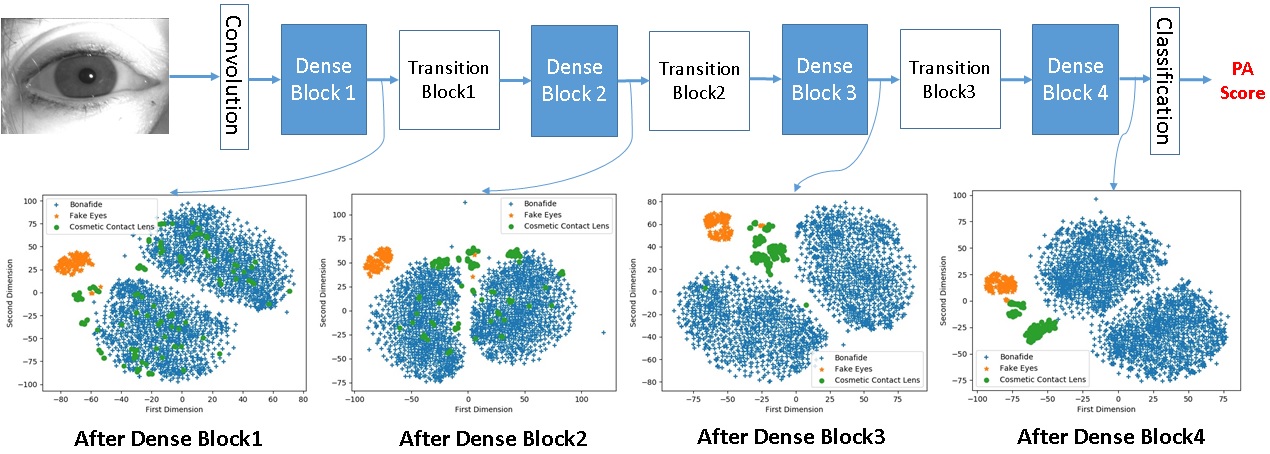}
	\caption{The architecture of D-NetPAD consists of four Dense blocks. We capture the features at the end of each Dense block, which are then visualized using t-sne plots (shown below each Dense block). The two-dimensional features of bonafide, artificial eyes, and cosmetic contacts overlap in the initial layers, but get separated in the last layer. The two blue clusters in each category correspond to the left and right eyes.}
	\label{fig:t-SNE-Plot}
\end{figure*}

We visualize the results of the D-NetPAD using t-Distributed Stochastic Neighbor Embedding (t-SNE) \cite{Maaten2008} plots and Gradient-weighted Class Activation Mapping (Grad-CAM) heatmaps. We utilize the D-NetPAD model trained on the training set of the Combined dataset for this purpose, and use the samples in the JHU-APL03 test set to generate these visualizations. The t-SNE helps in visualizing the features extracted from the D-NetPAD. It reduces the high-dimensional features extracted from the D-NetPAD to a lower dimension (two in our case), which are then used to construct a scatter plot. The architecture of the D-NetPAD consists of four Dense blocks. We capture the high-dimensional features at the end of each Dense block for visualization (Figure \ref{fig:t-SNE-Plot}). For instance, the feature set captured at the end of Dense block 4 has a size of $1 \times 1024 \times 7 \times 7$, which is flattened to $1 \times 50,176$. The 50,176-dimensional row vector is then reduced to a two-dimension vector. We draw three key observations from these plots:

\begin{enumerate}
	\item The distributions of bonafide, artificial eye and cosmetic contact features overlap after the initial Dense blocks, but separate for the later Dense blocks. As the depth of the network increases, the features of different categories are better separated. This substantiates the high performance of D-NetPAD (Table \ref{table:JHU-Results}).
	
	\item The features of different categories are sufficiently discriminated at the end of Dense Block 4, which justifies the use of four Dense blocks in the architecture as opposed to three in \cite{DYadav2019}.
	
	\item The plots shows two bonafide clusters which correspond to the left and right eyes. The left and right irides exhibit differences due to the orientation of upper and lower eyelids, location of specular reflection, the relative position of pupil center to iris center, and background illumination variation. The D-NetPAD captures these variations in its features. 
\end{enumerate}

\begin{figure}[h!]
	\includegraphics[width=\linewidth]{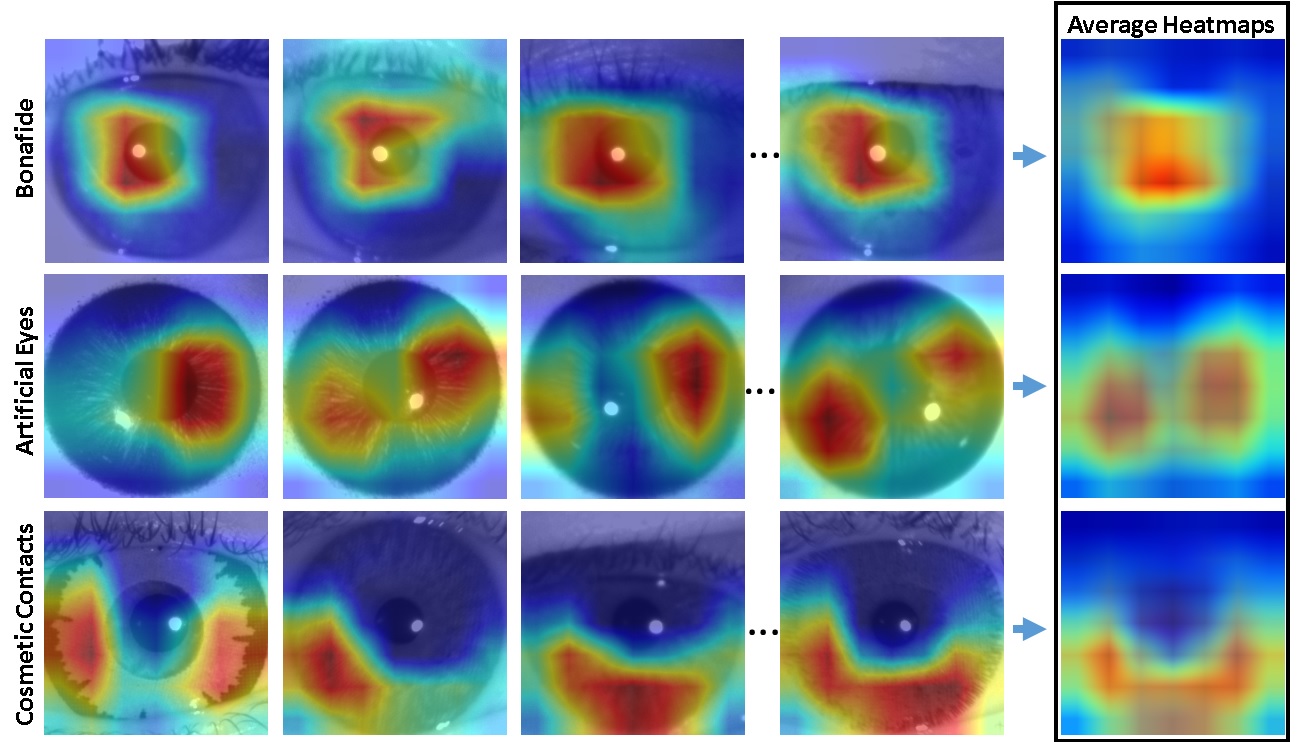}
	\caption{Grad-CAM \cite{Selvaraju2017} heatmaps corresponding to bonafide (first row), artificial eye (second row), and cosmetic contact (last row). The last column represents the average heatmaps of each category. The heatmaps represent focused regions of the image by the D-NetPAD algorithm. Red-colored regions represent highly focused regions by the D-NetPAD, whereas blue regions represent low priority ones.}
	\label{fig:Grad-CAM}
\end{figure}

We further visualize the CNN activations using the Grad-CAM \cite{Selvaraju2017} heatmaps. The Grad-CAM produces a coarse localization map highlighting the salient regions in an image that were used by the network to generate its inference. These are regions that produce high activations in the neural network. It is estimated using the gradient of a loss function, which backpropagates through the convolutional layers to the input image \cite{Selvaraju2017}. Figure \ref{fig:Grad-CAM} presents the CNN activation heatmaps on bonafide, artificial eye, and cosmetic contact images taken from the JHU-APL03 test set. The last column represents the average heatmaps of each category considering the entire test set. The red regions indicate high activation, whereas the blue regions represent low activation. The first row of Figure \ref{fig:Grad-CAM} shows the heatmap of bonafide sample images along with the average bonafide heatmap, where the high activation region is at the pupillary zone of the iris pattern. The second row of Figure \ref{fig:Grad-CAM} corresponds to the heatmap of artificial eye images, where the focus seems to be mainly on the left and right sub-regions of the iris. The last row shows the heatmaps of cosmetic contact images, where the lower sub-region of the iris pattern is focused. \textbf{The average heatmaps show the distinctive regions of focus in each category, which helps in discriminating bonafide from PAs}.  

\subsection{Spatial Frequency Analysis}
\begin{figure}[h!]
	\includegraphics[width=\linewidth]{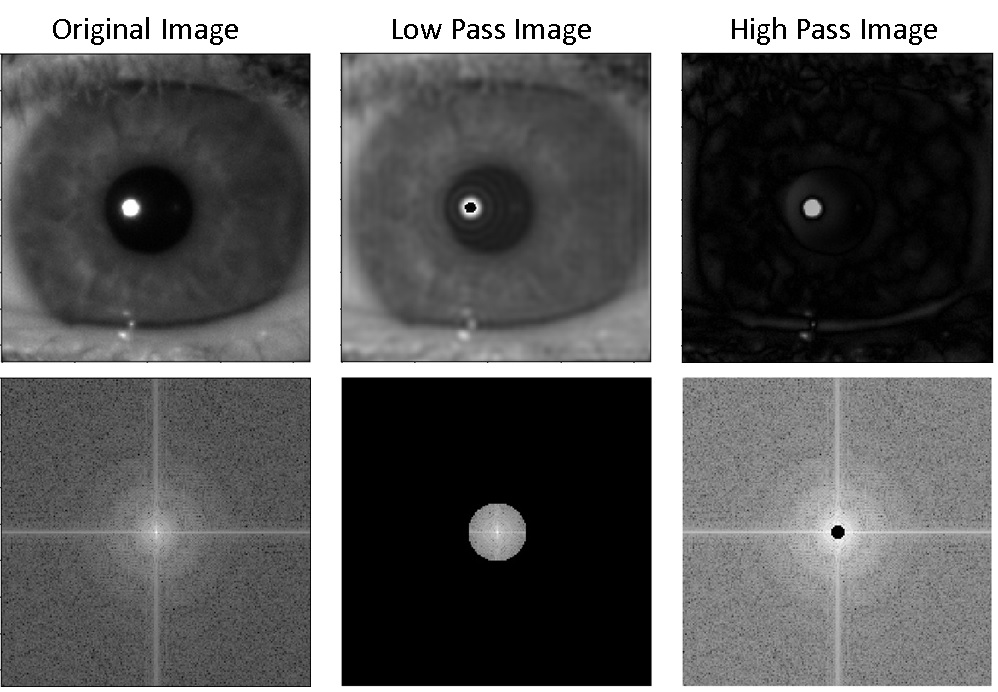}
	\caption{Frequency analysis of an input image. In the first row, the left-most image is the original image. The center image is a low-pass filtered image with a cutoff frequency of 20 (higher frequencies are suppressed), and the third is a high-pass filtered image with a cutoff frequency of 5 (lower frequencies are suppressed). The second row represents the corresponding fourier transforms.}
	\label{fig:FreqAnalysis}
\end{figure}

The iris is a highly textured pattern exhibiting numerous spatial frequencies. To understand what frequencies the D-NetPAD model has learned and how it impacts iris PAD performance, we perform a spatial frequency analysis on the D-NetPAD model. We attain the objective with the assumption that the performance of the model only gets affected by the manipulation of learned frequencies. We start by manipulating higher frequencies for two reasons. First, when we visually examine low- and high-pass filtered images (Figure \ref{fig:FreqAnalysis}), it is observed that a high-pass filter (suppression of low frequencies) considerably obscures the iris pattern. 
Second, deep learning-based models learn low frequencies first (initial epochs) and then high frequencies (later epochs) in the training process \cite{Rahaman2019, Xu2019}. In other words, the volume of weight parameters contributes towards expressing low frequencies is larger than the one expressing high frequencies \cite{Rahaman2019}. Due to this, small manipulation in low frequencies results in large shifts in the performance. In the case of high frequencies, more the architecture learns the high frequencies, more it tuned its parameters towards the learning delicacies of the training images, which may cause overfitting. So, learning of high frequencies determines the effectiveness of the model-fitting on the training data (i.e., efficiently fit or overfit).

\begin{figure}[h!]
	\centering
	\includegraphics[width=0.9\linewidth]{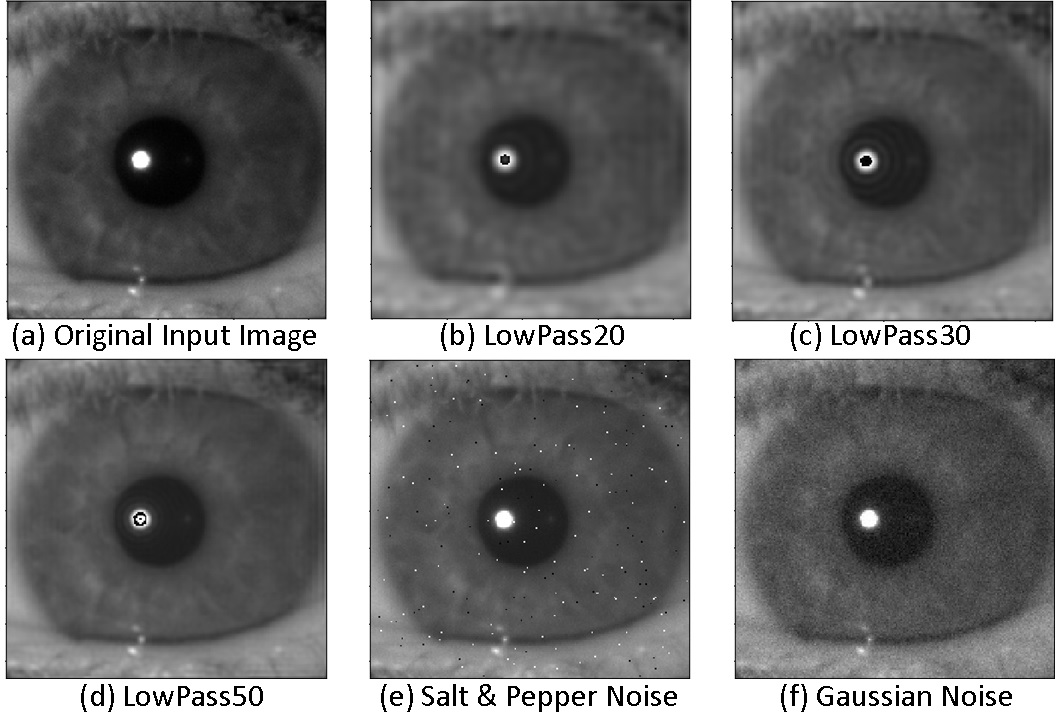}
	\caption{Different manipulations applied over the original input image (first image): low-pass filtered images with 20, 30, and 50 cutoff frequencies, salt and pepper noise, and Gaussian noise. Only test images are subject to these manipulations.}
	\label{fig:FreqAnalysisInput}
\end{figure}

\begin{figure}[h!]
	\centering
	\includegraphics[width=\linewidth]{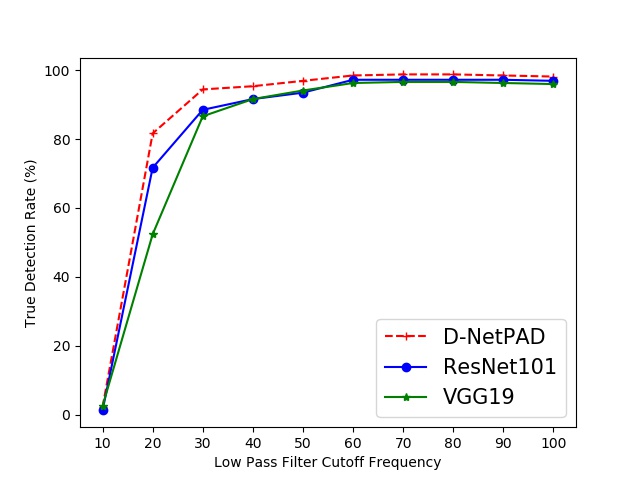}
	\caption{The plot of TDR (\%) @ 0.2\% FDR against different low-pass filter cutoff frequencies. Note the cutoff frequency beyond which the performance of D-NetPAD becomes stable (cutoff frequency 30 in this case). This cutoff frequency indicates that the D-NetPAD has not learned frequencies beyond this cutoff frequency. The performance steadiness of D-NetPAD is better than VGG19 and ResNet101.}
	\label{fig:LPFreq_Tolerance}
\end{figure}

For high frequency suppression, we use a low-pass filter with various cutoff frequencies. Cutoff frequency represents a radius from the center in the fourier transforms (second row of Figure \ref{fig:FreqAnalysis}). A low-pass filter allows frequencies below the cutoff frequency and attenuates higher frequencies. Figure \ref{fig:LPFreq_Tolerance} shows the performance of the D-NetPAD model along with the VGG19 and ResNet101 models on various low-pass filter cutoff frequencies. We use the train and test set of the Combined dataset for the experiments. The manipulation is only applied over the test images. There are two noteworthy observations. First, D-NetPAD shows a relatively lower drop in performance compared to VGG19 and ResNet101 models. Second, the performance of the D-NetPAD model becomes steady beyond the 30 cutoff frequency, which implies that the model has not overfitted to high frequencies beyond 30. Beyond a cutoff frequency of 60, the performance becomes constant implying that it has not learned any frequencies beyond 60. Another way of manipulating high frequencies is their addition to the input images, which we did by contaminating input images with salt and pepper noise. We also analyze the models when Gaussian noise (noise values are Gaussian-distributed) is added to the input images. Figure \ref{fig:FreqAnalysisInput} shows an example of an input image subject to high-frequency manipulation, (b) - (e), and the addition of Gaussian noise, (f). The performance is measured using a relative decrease in TDR (\%) at 0.2\% FDR. Table \ref{table:JHU-NoiseTolerance-Results} provides the results of VGG19, ResNet101, and D-NetPAD architectures when input images are manipulated. 

\textbf{The D-NetPAD model shows a lower decrease in TDRs compared to VGG19 and ResNet101 models when high frequencies in input images are manipulated (either suppression or addition).} The VGG19 and ResNet101 models have a large number of trainable parameters that result in the overfitting of these models to the training data. The overfitted models learn higher frequencies considerably well and, therefore, are more sensitive towards them. On the contrary, efficient learning of frequencies by the D-NetPAD makes it more robust towards manipulations to the high frequencies and also substantiates its generalizability across PAs, sensors, and datasets. Gaussian noise randomly affects both lower and higher frequencies, resulting in a higher drop in performance of all the networks, including D-NetPAD. 

\begin{table}[h!]
	\caption{Results in terms of TDR (\%) @ 0.2\% FDR and a relative decrease in TDR for VGG19, ResNet101, and D-NetPAD models, when high frequencies are manipulated (suppression or addition) or when Gaussian noise is applied to the input test images.}
	\label{table:JHU-NoiseTolerance-Results}
	\resizebox{\columnwidth}{!}{
	\begin{tabular}{|l|l|l|l|l|l|l|}
		\hline
		\multicolumn{1}{|c|}{\multirow{2}{*}{Input Test Images}} & \multicolumn{2}{c|}{VGG19} & \multicolumn{2}{c|}{ResNet101} & \multicolumn{2}{c|}{D-NetPAD} \\ \cline{2-7} 
		\multicolumn{1}{|c|}{} & \multicolumn{1}{c|}{\begin{tabular}[c]{@{}c@{}}TDR(\%)@ \\ 0.2\% FDR\end{tabular}} & \multicolumn{1}{c|}{\begin{tabular}[c]{@{}c@{}}Relative \\ Decreased \\ TDR (\%)\end{tabular}} & \multicolumn{1}{c|}{\begin{tabular}[c]{@{}c@{}}TDR(\%)@ \\ 0.2\% FDR\end{tabular}} & \multicolumn{1}{c|}{\begin{tabular}[c]{@{}c@{}}Relative \\ Decreased \\ TDR (\%)\end{tabular}} & \multicolumn{1}{c|}{\begin{tabular}[c]{@{}c@{}}TDR(\%)@ \\ 0.2\% FDR\end{tabular}} & \multicolumn{1}{c|}{\begin{tabular}[c]{@{}c@{}}Relative \\ Decreased \\ TDR (\%)\end{tabular}} \\ \hline
		Original Images & 96.26 & - & 96.88 & - & \textbf{98.58} & - \\ \hline
		\begin{tabular}[c]{@{}l@{}}LowPass20 \\ (Suppess high freq.)\end{tabular} & 52.33 & 45.63 & 71.65 & 26.04 & 81.61 & \textbf{17.21} \\ \hline
		\begin{tabular}[c]{@{}l@{}}LowPass30 \\ (Suppress high freq.)\end{tabular} & 86.60 & 10.03 & 88.47 & 8.68 & 94.39 & \textbf{4.25} \\ \hline
		\begin{tabular}[c]{@{}l@{}}LowPass50 \\ (Suppress high freq.)\end{tabular} & 94.08 & 2.26 & 93.45 & 3.54 & 96.88 & \textbf{1.72} \\ \hline
		\begin{tabular}[c]{@{}l@{}}Salt \& Pepper \\ (Add high freq.)\end{tabular} & 74.14 & 22.97 & 68.22 & 29.58 & 80.99 & \textbf{17.84} \\ \hline
		Gaussian Noise & 56.07 & 41.75 & 62.61 & \textbf{35.37} & 59.19 & 39.95 \\ \hline
	\end{tabular}
	}
\end{table}

\section{Conclusion}
We propose an effective and robust software-based iris PA detector called D-NetPAD. The D-NetPAD exploits the architectural benefits of DenseNet121. Experiments are performed on two datasets to help assess its effectiveness. The test sets of these datasets correspond to cross-PA, cross-sensor, and cross-dataset scenarios which measure the robustness of the D-NetPAD. We further explained the performance of the D-NetPAD using t-SNE plots, Grad-CAM heatmaps and frequency analysis. The source code and trained model are available at https://github.com/iPRoBe-lab/D-NetPAD.

\section*{Acknowledgment}
This research is based upon work supported in part by the Office of the Director of National Intelligence (ODNI), Intelligence Advanced Research Projects Activity (IARPA), via IARPA R\&D Contract No. 2017 - 17020200004. The views and conclusions contained herein are those of the authors and should not be interpreted as necessarily representing the official policies, either expressed or implied, of ODNI, IARPA, or the U.S. Government. The U.S. Government is authorized to reproduce and distribute reprints for governmental purposes notwithstanding any copyright annotation therein.

\balance
{\small
\bibliographystyle{ieee}
\bibliography{D-NetPAD_arXiv_V4}
}

\end{document}